\documentclass[sigconf]{acmart}

\AtBeginDocument{%
  }

\setcopyright{acmlicensed}

\copyrightyear{2018}
\acmYear{2018}
\acmDOI{XXXXXXX.XXXXXXX}

\acmConference[Conference acronym 'XX]{Make sure to enter the correct
  conference title from your rights confirmation email}{June 03--08}{Woodstock, NY}

\acmISBN{978-1-4503-XXXX-X/18/06}

\usepackage{multirow}
\usepackage{algorithmic}
\usepackage{algorithm}
\begin{document}

\title{Retrieve, Summarize, Plan: Advancing Multi-hop Question Answering with an Iterative Approach}


\author{Zhouyu Jiang}
\affiliation{%
  \institution{Ant Group}
  \city{Shanghai}
  \country{China}
}
\email{jiangzhouyu.jzy@antgroup.com}

\author{Mengshu Sun}
\affiliation{
  \institution{Ant Group}
  \city{Beijing}
  \country{China}
}
\email{mengshu.sms@antgroup.com}

\author{Zhiqiang Zhang}
\affiliation{
  \institution{Ant Group}
  \city{Hangzhou}
  \state{Zhejiang}
  \country{China}
}
\email{lingyao.zzq@antgroup.com}
\author{Lei Liang}
\affiliation{
  \institution{Ant Group}
  \city{Hangzhou}
  \state{Zhejiang}
  \country{China}
}
\email{leywar.liang@antgroup.com}






\renewcommand{\shortauthors}{Jiang et al.}

\begin{abstract}
Multi-hop question answering is a challenging task with distinct industrial relevance, and Retrieval-Augmented Generation (RAG) methods based on large language models (LLMs) have become a popular approach to tackle this task. Owing to the potential inability to retrieve all necessary information in a single iteration, a series of iterative RAG methods has been recently developed, showing significant performance improvements. However, existing methods still face two critical challenges: context overload resulting from multiple rounds of retrieval, and over-planning and repetitive planning due to the lack of a recorded retrieval trajectory. In this paper, we propose a novel iterative RAG method called \textbf{ReSP}, equipped with a dual-function summarizer. This summarizer compresses information from retrieved documents, targeting both the overarching question and the current sub-question concurrently. Experimental results on the multi-hop question-answering datasets HotpotQA and 2WikiMultihopQA demonstrate that our method significantly outperforms the state-of-the-art, and exhibits excellent robustness concerning context length.
\end{abstract}




\maketitle

\section{Introduction}
Open-domain question answering is a task that involves providing factual responses based on extensive documents \citep{voorhees-tice-2000-trec, zhang-etal-2023-survey-efficient} and is of significant application in hot industry scenarios such as intelligent assistants and generative search engines \citep{openai2024gpt4technicalreport, geminiteam2024geminifamilyhighlycapable}. Multi-hop question answering is one common and challenging sub-task within this field, requiring the system to integrate information to complete multi-step reasoning and answer questions \citep{mavi2024multihopquestionanswering}.

With the rapid development of large language models (LLMs), retrieval-augmented generation (RAG) based on these LLMs has become a popular method for addressing open-domain question answering \citep{siriwardhana-etal-2023-improving, 10.5555/3495724.3496517, 10.1145/3580305.3599931, 10.1145/3637528.3672065, qin-etal-2023-webcpm}. The typical RAG process involves using a retriever to recall documents from a corpus that are relevant to a given query and using these documents as context inputs for the LLMs to generate responses. However, when dealing with multi-hop question answering,  conventional RAG techniques frequently fall short of aggregating all the critical information within a singular retrieval iteration, leading to incomplete or incorrect answers. Consequently, a new genre of iterative RAG methods that incorporate question planning has recently been developed \citep{shao-etal-2023-enhancing, trivedi-etal-2023-interleaving, asai2024selfrag}. These methods assess after each retrieval whether the information at hand is sufficient for answering the question. If it is not, the methods generate a sub-question for the next step and perform another retrieval, iterating this process until the question can be satisfactorily answered. Owing to the employment of multiple retrieval iterations, iterative RAG has achieved a notable improvement in multi-hop question-answering scenarios compared to the single-round RAG approaches.

\begin{figure}[htbp]
    \centering
    \includegraphics[width=0.9\linewidth]{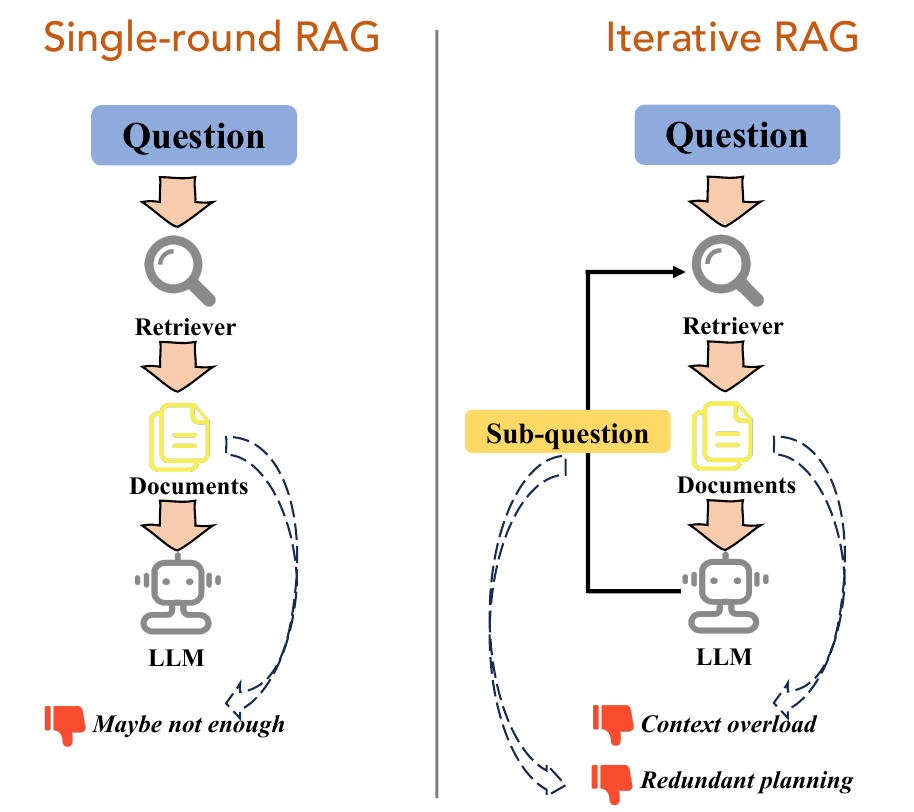}
    \caption{RAG pipelines illustration and challenges faced.}
    \label{issue}
\end{figure}
However, existing iterative RAG methods still encounter two principal challenges when handling multi-hop question answering. Firstly, due to multiple rounds of retrieval, iterative RAG methods have to deal with a longer context in contrast to single-round RAG methods, which consequently introduces more noise from the documents and increases the risk of the model missing key information during response generation \citep{liu-etal-2024-lost}. Secondly, current iterative RAG methods are heavily dependent on the model's interpretation of retrieved documents for decision-making, lacking a concrete record of the navigated trajectory. This makes it difficult for the model to discern whether the information needed to answer the overarching question has been sufficiently gathered and whether a sub-question has already been retrieved, leading to two issues: an over-planning scenario wherein the iterative process does not stop even despite sufficient information has been retrieved, and a repetitive planning scenario wherein a sub-question that has already been retrieved is reproduced \citep{yao2023react}.

The two challenges mentioned previously are primarily related to the effective processing of information obtained during the retrieval phase. To tackle this, drawing inspiration from query-focused summarization \citep{dang-2006-duc, xu-lapata-2020-coarse}, we introduce the \textbf{ReSP} (\textbf{Re}trieve, \textbf{S}ummarize, \textbf{P}lan) approach. This method not only condenses but also functionally decomposes the information accrued in each retrieval episode. Specifically, we integrate a novel LLM-based summarizer within the established iterative RAG framework and refine the iterative process. The summarizer undertakes dual roles: firstly, it compiles a summary of corroborative information from the retrieved documents for the overarching target question, termed the \textit{global evidence memory}; secondly, it crafts a response for the current sub-question based on the retrieved documents, termed the \textit{local pathway memory}. At the inception of each iteration, the accumulated global evidence memory and local pathway memory are combined as contextual inputs for the model's evaluation. Should the information be evaluated as adequate, the procedure advances to the generation of the final response; otherwise, a new sub-question is formulated, with the requirement that the model must not generate previously retrieved sub-questions.

Our experimental findings reveal that, under uniform experimental settings, the ReSP model markedly surpasses a range of current single-round and iterative RAG approaches when evaluated on multi-hop question-answering benchmarks such as HotpotQA \citep{yang-etal-2018-hotpotqa} and 2WikiMultihopQA \citep{ho-etal-2020-constructing}. Notably, it exhibits a substantial enhancement in performance, with an increase of 4.1 F1 score over the state-of-the-art (SOTA) on HotpotQA, and an improvement of 4.4 F1 score over the SOTA on 2WikiMultihopQA. Furthermore, we conducted a series of in-depth comparative studies to discern the effect of base model choice on its performance and confirmed that ReSP possesses commendable robustness to context length compared to other RAG methods.

In conclusion, our contributions are as follows:
\begin{itemize}
    \item 
    Targeting the multi-hop question-answering scenario, we propose an innovative iterative RAG approach that incorporates query-focused summarization to tackle the context overload problem resulting from multiple rounds of retrieval. In particular, we have refined the summarizer's function to compress information aimed at both the overarching question and the current sub-question, thereby optimizing issues related to over-planning and repetitive planning.
    \item 
    Experimental results show that our approach significantly surpasses existing methods in performance, and it exhibits considerable robustness to variations in context length.
\end{itemize}

\section{Related Works}

\textbf{Retrieval-Augmented Generation.} Retrieval-Augmented Generation (RAG) enhances LLMs by retrieving relevant documents from external databases and integrating them into the generation process \citep{10.5555/3495724.3496517, Khandelwal2020Generalization, izacard-grave-2021-leveraging, jiang-etal-2024-efficient, zhou2024metacognitive}. Recent work can be divided into single-round RAG \citep{kim2024sure, xu2024recomp, shi-etal-2024-replug} and iterative RAG \citep{trivedi-etal-2023-interleaving, shao-etal-2023-enhancing, jiang-etal-2023-active, asai2024selfrag} based on the number of retrieval rounds. In multi-hop question-answering scenarios, iterative RAG often achieves better results because it allows for detailed decomposition of the question. However, due to the increased number of iterations, iterative RAG faces challenges in long-context processing.

\textbf{Query-focused summarization.} 
Query-focused summarization (QFS) aims to extract or generate a summary of an input document that directly answers or is relevant to a given query\citep{dang-2006-duc}. Due to the higher requirements of QFS tasks regarding annotation accuracy and data diversity compared to typical summarization tasks, data scarcity has long-limited performance improvement. Previous work has mainly focused on dataset construction \citep{nema-etal-2017-diversity, baumel2016topic, zhong2021qmsum, xu-etal-2023-lmgqs}, proposing specialized datasets for scenarios such as news, debates, and meetings.

Recent work has attempted to introduce QFS into single-round RAG for refined information compression \citep{xu2024recomp, edge2024localglobalgraphrag}but has not extended to iterative RAG, nor has it decomposed the functions of QFS. To the best of our knowledge, we are the first to integrate QFS with iterative RAG to address multi-hop question answering. 
\begin{figure*}
    \centering
    \includegraphics[width=0.85\linewidth]{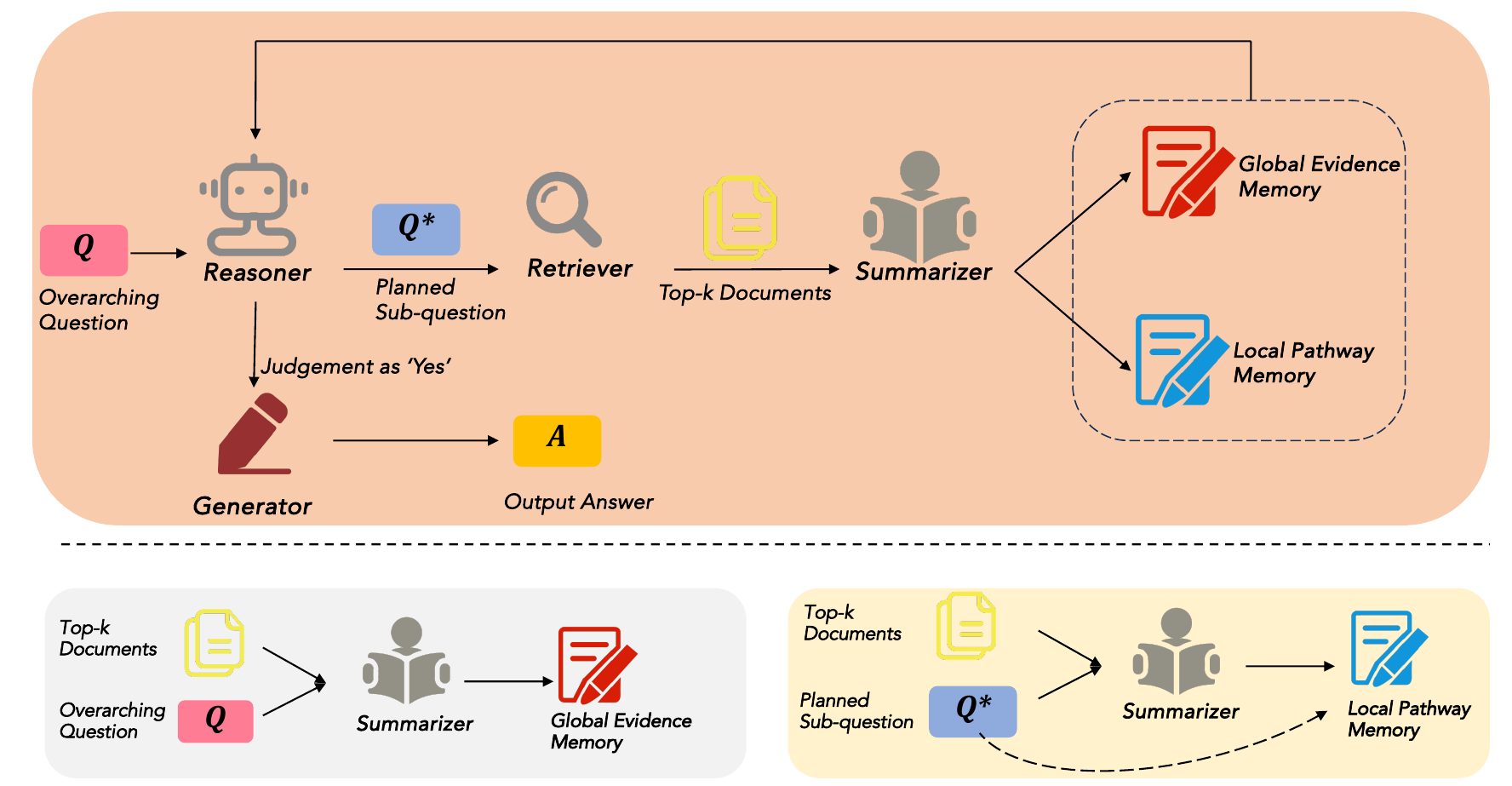}
    \caption{The ReSP framework consists of four modules: Reasoner, Retriever, Summarizer, and Generator. The reasoner makes decisions based on the current memory queues, determining whether to exit the iteration and generate a response or to produce a sub-question for further iteration. The retriever searches the corpus based on the sub-question provided by the reasoner (for the first iteration, the sub-question is the same as the overarching question, thus the reasonser is bypassed). The summarizer performs dual summarization on the retrieved documents, extracting information relevant to both the overarching question \textit{Q} and the current sub-question \textit{Q*}, and stores the summaries in the global evidence memory and local pathway memory queues respectively. The generator produces answer \textit{A} based on the information in the memory queues. }
    \label{fig-resp}
\end{figure*}

\section{Preliminary}
In this section, we formulate the task of single-round RAG and iterative RAG.
\subsection{Single-round RAG}
Given a question $q$ and a retrieval corpus $D=\{{d_i\}}_{i=1}^{|D|}$, the retrieval-augmented LLM needs to generate an answer $y$ based on the retrieved documents relevant to the question. The process can be represented as:
\begin{equation}
     y = {\rm LLM_{Gen}}([q, D_q], \rm Prompt_{Gen})   
\end{equation}
where $D_q$ is the retrieved documents for the question $q$. $\rm Prompt_{Gen}$ is the designated prompt for QA task. $\rm LLM_{Gen}$ is the role of LLM, which focuses on QA task.
\subsection{Iterative RAG}
Compared to single-round RAG, iterative RAG introduces an evaluation and decision-making process for the current state in each iteration. Specifically, in each iteration, the LLM will provide an action $a$, which has only binary values $\rm (exit, continue)$, and can be represented as:
\begin{equation}
    a = {\rm LLM_{judge}}([q, {state}], \rm Prompt_{judge})
\end{equation}
where $state$ refers to the historical retrieved documents or the information derived from them. $\rm Prompt_{judge}$ is the designated prompt for judgment. $\rm LLM_{judge}$ is an LLM used to assess the state and provide action signals.

The next step is determined based on the value of $a$:
\begin{equation}
    next\_step = \begin{cases}
        {\rm LLM_{plan}}([q, {state}], {\rm Prompt_{plan}}), a={\rm continue} \\
        {\rm answer\_generation}, a={\rm exit}
    \end{cases}
\end{equation}
where $\rm Prompt_{plan}$ is the designated prompt for sub-question generation. $\rm LLM_{plan}$ is an LLM for breaking down problems.
\section{Methodology}
Figure \ref{fig-resp} illustrates our ReSP framework, which consists of four components: \texttt{Reasoner}, \texttt{Retriever}, \texttt{Summarizer}, and \texttt{Generator}. Essentially, Reasoner, Summarizer, and Generator are all based on a fine-tuning-free LLM, designed to execute specific tasks using an array of carefully selected prompt engineering techniques. For specific prompt templates, please see Table \ref{tab:prompt}. Our main contribution lies in the Summarizer, while the design of the other modules is similar to that of conventional iterative RAG methods.

\begin{table*}[]
\caption{The prompt templates of different modules in ReSP. The input parameters include \texttt{Overarching question}: the input question; \texttt{Sub-question}: the current sub-question generated by the reasoner; \texttt{Combined memory queues}: the concatenated content of the global evidence memory and local pathway memory queues for each iteration; \texttt{docs}: the documents retrieved from a single round of retrieval.}
\begin{tabular}{@{}ccp{10cm}@{}}
\toprule
\textbf{Module}                      & \textbf{Function}            & \textbf{Prompt}                                                                                                                                                                                                                                                                                                                                                                                                                                                                                                                                                                                                                                             \\ \midrule
\multirow{10}{*}{Reasoner}   & \multirow{5}{*}{Judge}               & Judging based solely on the current known information and without allowing for inference, are you able to completely and accurately respond to the question \texttt{Overarching question}? \texttt{\textbackslash{}n}Known information: \texttt{Combined memory queues}. \texttt{\textbackslash{}n}If you can, please reply with "Yes" directly; if you cannot and need more information, please reply with "No" directly.                                                                                                                                                                                                                                                             \\
                            & \multirow{5}{*}{Plan}                & You serve as an intelligent assistant, adept at facilitating users through complex, multi-hop reasoning across multiple documents. Please understand the information gap between the currently known information and the target problem.Your task is to generate one thought in the form of question for next retrieval step directly. DON'T generate the whole thoughts at once!\texttt{\textbackslash{}n} DON'T generate thought which has been retrieved.\texttt{\textbackslash{}n} {[}Known information{]}: \texttt{Combined memory queues}\texttt{\textbackslash{}n}{[}Target question{]}: \texttt{Overarching question}\texttt{\textbackslash{}n}{[}You Thought{]}: \\
\hline
\multirow{10}{*}{Summarizer} & \multirow{5}{*}{Global Evidence}     & Passages: \texttt{docs}\texttt{\textbackslash{}n}Your job is to act as a professional writer. You will write a good-quality passage that can support the given prediction about the question only based on the information in the provided supporting passages. Now, let's start. After you write, please write {[}DONE{]} to indicate you are done. Do not write a prefix (e.g., "Response:") while writing a passage.\texttt{\textbackslash{}n}Question:\texttt{Overarching question}\texttt{\textbackslash{}n}Passage:                                                                                                                                                                      \\
                            & \multirow{5}{*}{Local Pathway}       & Judging based solely on the current known information and without allowing for inference, are you able to respond completely and accurately to the question \texttt{Sub-question}? \texttt{\textbackslash{}n}Known information: \texttt{Combined memory queues}. If yes, please reply with "Yes", followed by an accurate response to the question \texttt{Sub-question}, without restating the question; if no, please reply with "No" directly.                                                                                                                                                                                                                                      \\
\hline
\multirow{4}{*}{Generator}                   & \multirow{4}{*}{Response Generation} & Answer the question based on the given reference.\texttt{\textbackslash{}n}Only give me the answer and do not output any other words.\texttt{\textbackslash{}n}The following are given reference: \texttt{Combined memory queues}\texttt{\textbackslash{}n}Question: \texttt{Overarching question}                                                                                                                                                                                                                                                                                                                                                                                            \\ \bottomrule
\end{tabular}

\label{tab:prompt}
\end{table*}

\subsection{Dual-Function Summarizer}
As mentioned earlier, our goal is to address issues of context overload and redundant planning. To tackle context overload, a straightforward approach is to employ summarization to compress information. However, even with summarization, the model still lacks an explicit record of the planning path, which does not resolve the issue of redundant planning. During the iterative process, over-planning could arise if summaries overlook information crucial for directly addressing the overarching question, or repetitive planning might occur if the information difference between different rounds of summaries is not significant. Therefore, a more sophisticated design of the summarizer is necessary to distinguish the functions of various pieces of information.

Drawing on the idea of query-focused summarization, we have designed a dual-function summarizer. Confronted with retrieved documents, this summarizer concurrently executes two tasks: producing summaries of supporting information pertinent to the overarching question and generating responses for the current sub-question, while managing two distinct memory queues--the global evidence memory and the local pathway memory. Summaries related to the overarching question are deposited into the global evidence memory, serving to explicitly signal the model to cease iterations when information is enough, thus mitigating the risk of over-planning. Concurrently, responses for the current sub-question, alongside the sub-question itself, are stored in the local pathway memory. This explicitly guides the model's recognition of the progress in the question planning path as well as the sub-questions that have been historically retrieved, preventing repetitive planning.
\subsection{Summary-Enhanced Iterative RAG Process}
Here we delineate the refined iterative RAG workflow that incorporates the dual-function summarizer. The pseudocode for the algorithm implementation is shown as Algorithm \ref{alg_resp}.

Given a query \textit{Q} and a document corpus \textit{D}, we initially deploy a retriever to identify the \textit{K} documents from \textit{D} that are most relative to \textit{Q}. These documents are then directed into the summarizer for summary creation and memory queue updates (note that during the first iteration, the sub-question is essentially the overarching question, so there is no generation of response for the current sub-question). Subsequently, the contents of the two memory queues are concatenated to provide context input for the reasoner, which is responsible for determining whether the current information is sufficient to address the overarching question. Should it be adequate, the iterative process is halted, and the memory queues are utilized as context for the generator to produce the final answer. If the information is insufficient, the reasoner generates a subsequent sub-question \textit{Q*} that is distinct from previously retrieved sub-questions based on the current context, prompting the next iteration round.
\begin{algorithm}
    \caption{Multi-hop QA through ReSP}
    \label{alg_resp}
    \renewcommand{\algorithmicrequire}{\textbf{Input:}}
    \renewcommand{\algorithmicensure}{\textbf{Output:}}
    \begin{algorithmic}
        \REQUIRE Overarching question $Q$, Document corpus $D$, Retriever $R$, LLMs with different roles  $(LLM_{Reason}, LLM_{Sum}, LLM_{Gen})$, Max number of iterations $M$
        \STATE $if\_finished \gets {\rm False}$
        \STATE $run\_cnt \gets 0$
        \STATE $q \gets Q$
        \STATE $S_{Global} \gets []$
        \STATE $S_{Local} \gets []$
        \WHILE{not $if\_finished$ and $run\_cnt < M$}
            \STATE $run\_cnt \gets run\_cnt + 1$
            \STATE $D_q \gets R(q)$
            \IF{$q \neq Q$}
                \STATE $S_{Local} \gets S_{Local} + LLM_{Sum}([q, D_q], {\rm Prompt_{local}})$
            \ENDIF
            \STATE $S_{Global} \gets S_{Global} + LLM_{sum}([Q, D_q], {\rm Prompt_{global}})$
            \STATE $if\_finished \gets LLM_{Reason}([Q, S_{Global}, S_{Local}], {\rm Prompt_{judge}})$
            \IF{not $if\_finished$}
                \STATE $Q^* \gets LLM_{Reason}([Q, S_{Global}, S_{Local}], {\rm Prompt_{plan}})$
                \STATE $q \gets Q^*$
            \ENDIF
        \ENDWHILE
        \STATE $A \gets LLM_{Gen}([Q, S_{Global}, S_{Local}], {\rm Prompt_{Gen}})$
        \ENSURE Final answer $A$

    \end{algorithmic}
        
\end{algorithm}

\section{Experimental Settings and Results}
\subsection{Datasets}
We conduct experiments on two multi-hop question-answering benchmark datasets: HotpotQA \citep{yang-etal-2018-hotpotqa} and 2WikiMultihopQA \citep{ho-etal-2020-constructing}. These two datasets are both constructed based on Wikipedia documents, which enables the utilization of a consistent document corpus and retrievers to provide external references for LLMs. Following the open-sourced RAG toolkit FlashRAG \citep{FlashRAG}, we employ its preprocessed dataset format. For each dataset, we utilize the first 1,000 entries from the original development set for testing. We report the token-level F1 score of answer strings to evaluate the quality of the generation.
\subsection{Experimental Setup}
In our main experiments, we employ the Llama3-8B-instruct\citep{llama3modelcard} as the base model and E5-base-v2 \citep{wang2022text} as the retriever. The retrieval corpus comprises Wikipedia data from December 2018, consisting of approximately 20 million pre-segmented passages. For the model and data links, please refer to the FlashRAG open-source repository \footnote{https://github.com/RUC-NLPIR/FlashRAG}.

The model's maximum input length is set to 12,000, and the maximum output length is set to 200. For each query, we retrieve the top-5 documents from the corpus based on vector similarity as the result. The maximum number of iterations is set to 3. If the retrieval process is still in iteration after 3 attempts, the model will directly proceed to the final response generation. All experiments are conducted on 4 NVIDIA A100 GPUs.
\subsection{Baselines}
We select representative methods from single-round RAG and iterative RAG as baselines for comparison. 

\textbf{Direct Prompting:} LLM directly generates answers without retrieval.

{\textbf{Single-round RAG:}} \textbf{Standard RAG} directly generates answers based on all retrieved documents.
\textbf{SuRe} \citep{kim2024sure} constructs and ranks summaries of the retrieved passages for each of the multiple answer candidates.
\textbf{RECOMP} \citep{xu2024recomp} compresses retrieved documents into textual summaries before in-context integration.
\textbf{REPLUG} \citep{shi-etal-2024-replug} prepends each retrieved document separately to the input context and ensembles output probabilities from different passes.

{\textbf{Iterative RAG:}} \textbf{Iter-RetGen} \citep{shao-etal-2023-enhancing} leverages the model output from the previous iteration as a specific context to help retrieve more relevant knowledge.
\textbf{Self-RAG} \citep{asai2024selfrag} enhances an LM's quality and factuality through retrieval and self-reflection.
\textbf{FLARE} \citep{jiang-etal-2023-active} iteratively uses the upcoming sentence to anticipate future content, which is
then utilized as a query to retrieve relevant documents to regenerate the sentence if it contains low-confidence tokens.
\textbf{IRCoT} \citep{trivedi-etal-2023-interleaving} guides the retrieval with Chain-of-Thought (CoT) \citep{wei2022chain} and in turn uses retrieved results to improve CoT.

\subsection{Main Results}
\label{sec:main}
\begin{table*}[htbp]
\caption{Performance comparison on HotpotQA and 2WikiMultihopQA. All methods except Self-RAG utilize fine-tuning-free Llama3-8B-instruct for generation. Self-RAG uses trained selfrag-llama2-7B released by authores. We run experiments five times and report the average token-level F1 score of answer strings. The result of ReSP outperforms baseline models in t-test at \textit{p} < 0.05 level. The best results are in bold and the second best results are underlined.}
\begin{tabular}{@{}cccc@{}}
\toprule
Method       & Pipeline type & HotpotQA & 2WikiMultihopQA \\ \midrule
Direct Prompting & --          &28.4      &33.9   \\
Standard RAG & Single-round RAG   & 38.6     & 20.1  \\
SuRe \citep{kim2024sure}        & Single-round RAG   & 33.4     & 20.6  \\
RECOMP\citep{xu2024recomp}       & Single-round RAG   & 37.5     & 32.4  \\
REPLUG\citep{shi-etal-2024-replug}       & Single-round RAG   & 31.2     & 21.1  \\
Iter-RetGen\citep{shao-etal-2023-enhancing}  & Iterative RAG& 38.3     & 21.6  \\
Self-RAG\citep{asai2024selfrag}  & Iterative RAG& 29.6     & 25.1  \\
FLARE \citep{jiang-etal-2023-active} & Iterative RAG & 28.0     & \underline {33.9}  \\
IRCoT\citep{trivedi-etal-2023-interleaving}        & Iterative RAG & \underline{43.1}     & 32.4  \\ \midrule
ReSP(ours)   & Iterative RAG & \textbf{47.2}     & \textbf{38.3}  \\ \bottomrule
\end{tabular}

\label{tab:main_exp}
\end{table*}

Our results on HotpotQA and 2WikiMultihopQA are displayed in Table \ref{tab:main_exp}. First, we notice that iterative RAG, especially IRCoT, demonstrates significant performance improvements compared to single-round RAG. This suggests that conducting multiple rounds of retrieval can indeed capture information more comprehensively and produce more accurate responses. Second, within single-round RAG, RECOMP, which incorporates summarization, exhibits superior performance, indicating that summarization is an effective method of information compression even within single-round RAG. These findings validate the rationale behind our approach, which combines multi-round retrieval with summarization.

Our method, ReSP, achieves significant improvements on both datasets, outperforming the SOTA by \textbf{4.1} F1 score on HotpotQA and \textbf{4.4} F1 score on 2WikiMultihopQA, surpassing a range of existing iterative RAG methods. This demonstrates the effectiveness of the approach we propose.

To analyze whether ReSP can address context overload and redundant planning issues, we record key statistics during the model's execution.

Initially, concerning the issue of context overload, utilizing HotpotQA as an example, when the maximum number of iterations is limited to three, the recorded average total word count of the retrieved documents during the process is 509.59, while the average word count of the evidence information provided to the generator is 86.61, culminating in a compression ratio of 5.88. This indicates that ReSP significantly compresses useful information during the process, helping to prevent context overload.

Regarding redundant planning, ReSP exhibits an average iteration count of 1.24 for HotpotQA and 1.60 for 2WikiMultihopQA, while IRCoT has counts of 1.80 and 2.32, respectively. This indicates that ReSP considerably minimizes planning steps.

Additionally, as shown in Table \ref{tab:time}, we provide a comparison of computation time. We record the runtime of standard RAG, IRCoT, and ReSP on 200 instances, maintaining the same experimental environment described before. The results indicate that the iterative ReSP takes longer compared to the single-round standard RAG, which may make it less suitable for time-sensitive scenarios unless specialized engineering optimizations are performed. However, compared to IRCoT, which is also an iterative method, ReSP significantly reduces runtime by compressing context and minimizing redundant planning while achieving better performance.

It is noteworthy that ReSP allows for the exposure of both global and local evidence information collected at each step of the problem-solving process. This is particularly suitable for scenarios that demand high transparency and trustworthiness of model outputs, such as legal and medical fields. In these contexts, users can understand how the problem is decomposed, how information is retrieved, and how the final answer is organized during the process. This advantage is not present in standard RAG and other iterative RAG methods that do not record the retrieval process.

\begin{table}[]
\centering
\caption{Comparison of
computation time.}
\begin{tabular}{@{}ccc@{}}
\toprule
Method       & HotpotQA & 2WikiMultihopQA \\ \midrule
Standard RAG & 1.1min   & 1.2min          \\
IRCoT        & 41.8min  & 48.2min         \\
ReSP         & 9.5min   & 13.7min         \\ \bottomrule
\end{tabular}
\label{tab:time}
\end{table}

\begin{table*}[htbp]
\centering
\caption{Impact of base model size on different modules.}
\begin{tabular}{@{}ccccc@{}}
\toprule
\texttt{Reasoner}            & \texttt{Summarizer}          & \texttt{Generator}          & HotpotQA & 2WikiMultihopQA \\ \midrule
Llama3-8B-Instruct  & Llama3-8B-Instruct  & Llama3-8B-Instruct & 47.2     & 38.3            \\
\textbf{Llama3-70B-Instruct} & Llama3-8B-Instruct  & Llama3-8B-Instruct & 48.8(\textcolor{red}{+1.6pt})     & 37.2(\textcolor{blue}{-1.1pt})            \\
Llama3-8B-Instruct  & \textbf{Llama3-70B-Instruct} & Llama3-8B-Instruct & 47.3(\textcolor{red}{+0.1pt})     & 34.1(\textcolor{blue}{-4.2pt})            \\
Llama3-8B-Instruct  & Llama3-8B-Instruct  & \textbf{Llama3-70B-Instruct} & 48.2(\textcolor{red}{+1.0pt})     & 38.7(\textcolor{red}{+0.4pt})            \\ \bottomrule
\end{tabular}

\label{tab:module_exp}
\end{table*}

\section{Empirical Analysis}
To further analyze ReSP, we conduct comparative experiments to investigate the adaptability across different open-source base models and the impact of the base model size on modules' performance. Additionally, we examine the method's robustness concerning two key variables: context length and maximum number of iterations. Through case studies, we demonstrate that ReSP can mitigate the issues of over-planning and repetitive planning.
\subsection{Adaptability across different base models}
Considering popularity, performance, and computational cost, we utilized the open-source Llama3-8B-Instruct as the base model in our main experiments to validate the effectiveness of our method. In practical scenarios, there are many other base model series to choose from. To assess the adaptability of our method across different base model series, we select three open-source base models of similar size to Llama3-8B-Instruct for experiments, maintaining consistency in other experimental settings: \textbf{Qwen2.5-7B-Instruct}\citep{qwen2.5}, \textbf{gemma-2-9b-it}\citep{Gemma}, and \textbf{Deepseek-V2-Lite-Chat}\citep{deepseekv2}. All models can be available on HuggingFace\footnote{https://huggingface.co/Qwen/Qwen2.5-7B-Instruct}\footnote{https://huggingface.co/google/gemma-2-9b-it}\footnote{https://huggingface.co/deepseek-ai/DeepSeek-V2-Lite-Chat}.

\begin{figure}[htbp]
    \centering
    \includegraphics[width=0.85\linewidth]{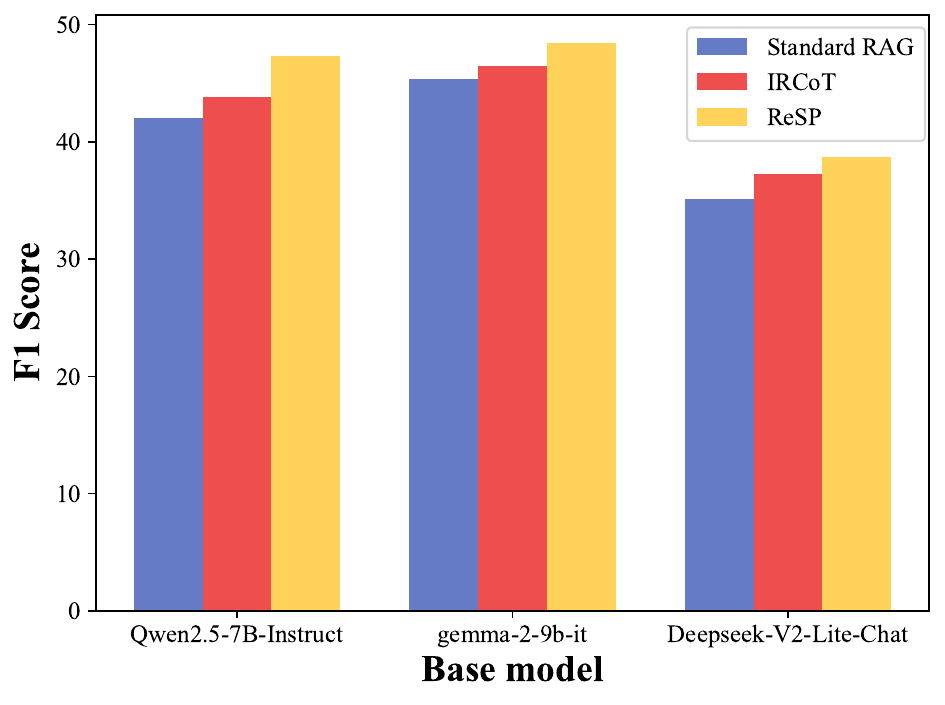}
    \caption{Performance comparison across different base models. We report the token-level F1 score obtained from testing on the HotpotQA dataset.}
    \label{fig-adapt}
\end{figure}

The results on HotpotQA are shown in Figure \ref{fig-adapt}. It can be observed that, compared to Standard RAG and IRCoT, the ReSP method consistently achieves significant improvements across three different base models. This demonstrates our method's strong adaptability to various base models.
\subsection{Impact of the base model size}

ReSP features a modular design, wherein each module works independently, allowing for the use of different base models to collaborate. To provide empirical guidance for model selection in practical applications, we test how the size of the model affects the performance of each module.

We substitute Llama3-70B-Instruct for Llama3-8B-Instruct and use this larger model as the base for the Reasoner, Summarizer, and Generator modules respectively, comparing the effect changes with the original results. Table \ref{tab:module_exp} presents our experimental results.

Firstly, regarding the reasoner module, the changes are inconsistent across the two datasets, with an improvement on HotpotQA but a decline on 2WikiMultihopQA. The reason for this inconsistency is that 2WikiMultihopQA has questions with more logical hops compared to HotpotQA. A larger model is likely to give more detailed planning steps, leading to a failure to obtain all the necessary information to answer the question within the set maximum number of iterations, hence causing a drop in performance.

Secondly, for the summarizer module, we observe that using a larger model size does not result in performance improvements; in fact, there is a significant decline on 2WikiMultihopQA. Upon reviewing the summarization logs, we find that Llama3-70B-Instruct is more lenient in discerning relevance. It tends to extract information that seems related but is actually irrelevant to the question, which can disrupt the planning and ultimately the generation of responses.

Lastly, regarding the generator module, we observe consistent improvements when using Llama3-70B-Instruct, which suggests that even when provided with clear evidence, a model with stronger semantic comprehension capabilities still aids in generating more accurate responses.

In summary, in real-world applications, it is advisable to allocate a larger base model to the generator, as long as the available resources allow for it. However, for the reasoner module, if the allowable number of iterations is low, there is no need to use a larger base model. The summarizer also does not require a larger base model.

\subsection{Robustness to context length}
\begin{figure}[htbp]
    \centering
    \includegraphics[width=0.9\linewidth]{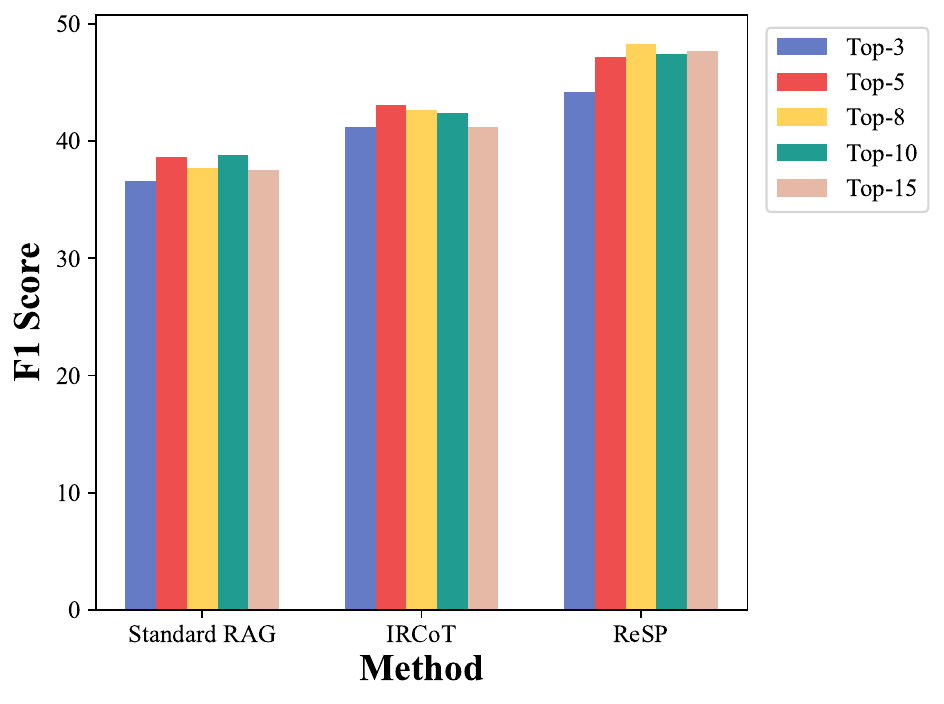}
    \caption{Bar chart of the performance variations of different RAG methods with varying numbers of retrieved documents per iteration. We report the token-level F1 score obtained from testing on the HotpotQA dataset.}
    \label{fig-robust}
\end{figure}
To further determine whether ReSP can address the issue of context overload, we adjust the number of documents retrieved per iteration and observe the changes in performance. Our comparison involves using standard RAG and IRCoT as control groups. These experiments are carried out on the HotpotQA dataset. Results are shown in Figure \ref{fig-robust}.
Firstly, when the number of documents retrieved per iteration \textit{k} is set to 3, all three methods experience a notable decrease in performance. This indicates that the adequacy of information retrieved in a single round significantly affects both single-round and iterative RAG methods. Therefore, it is necessary to extend the context length during application appropriately.

When \textit{k} is greater than 5, the standard RAG and IRCoT exhibit a performance degradation trend. Particularly, IRCoT, which utilizes iterative retrieval, suffers from a more significant performance drop due to the accumulation of retrieved information. This demonstrates that context overload has a pervasive impact on existing RAG methods.

Our method demonstrates exceptional robustness to context length, delivering consistent performance regardless of whether \textit{k} is set to 15 or 5. This is because we extract key information in each iteration, effectively maintaining a stable and concise context for the generator. As a result, the generator remains unaffected by changes in the length of retrieved documents during the process.
\begin{table*}[htbp]
\caption{Case studies comparing IRCoT and ReSP. They are conducted to provide evidence of ReSP's capability in addressing the issues of over-planning and repetitive planning.}
\begin{tabular}{@{}lp{3cm}ll@{}}
\toprule
Issue               & Question                                                                                     & IRCoT                                                                                                                                                                                                                                                                                                                      & ReSP                                                                                                                                                                                                                                                                                                      \\ \midrule
Over-planning       & \begin{tabular}[c]{@{}p{3cm}@{}}Victor Varnada directed Twisted Fortune, which starred which brother of Eddie Murphy? \end{tabular}       & \begin{tabular}[c]{@{}p{4.5cm}@{}}\textbf{Round 0 Thought:} \\ "The text does not mention Twisted Fortune, it talks about Eddie Murphy's career and his brother Charlie Murphy. What's the next step?"\end{tabular} & \begin{tabular}[c]{@{}p{4.5cm}@{}}\textbf{Round 0 Thought:}\\ {[}Global Evidence Memory{]}: "Victor Varnado directed the black comedy Twisted Fortune, which starred Charlie Murphy, a brother of Eddie Murphy."\\ --\textgreater{}{[}Judgment{]}: "Yes"\\ --\textgreater{}{[}Response{]}: "Charlie Murphy"\end{tabular} \\
\midrule
Repetitive planning & \begin{tabular}[c]{@{}p{3cm}@{}}What is Rachelle Amy Beinart's role in the film that follows a group of rebels on a mission?\end{tabular} & \begin{tabular}[c]{@{}p{4.5cm}@{}}\textbf{Round 0 Thought:}\\ "Rachelle Amy Beinart is not mentioned in any of the provided Wikipedia articles."\\ \textbf{Round 1 Thought:}\\ "Rachelle Amy Beinart is not mentioned."\\ ...\end{tabular}                                                                                                          & \begin{tabular}[c]{@{}p{4.5cm}@{}}\textbf{Round 0 Thought:}\\ {[}Sub-question{]}: "Who is Rachelle Amy Beinart?"\\ \textbf{Round 1 Thought:}\\ {[}Sub-question{]}: "What film mentioned in the provided passages features a group of rebels on a mission?"\\ ...\end{tabular}                                                      \\ \bottomrule
\end{tabular}

\label{tab:case_study}
\end{table*}
\subsection{Exploration of the number of iterations}
\begin{figure}[htbp]
    \centering
    \includegraphics[width=0.8\linewidth]{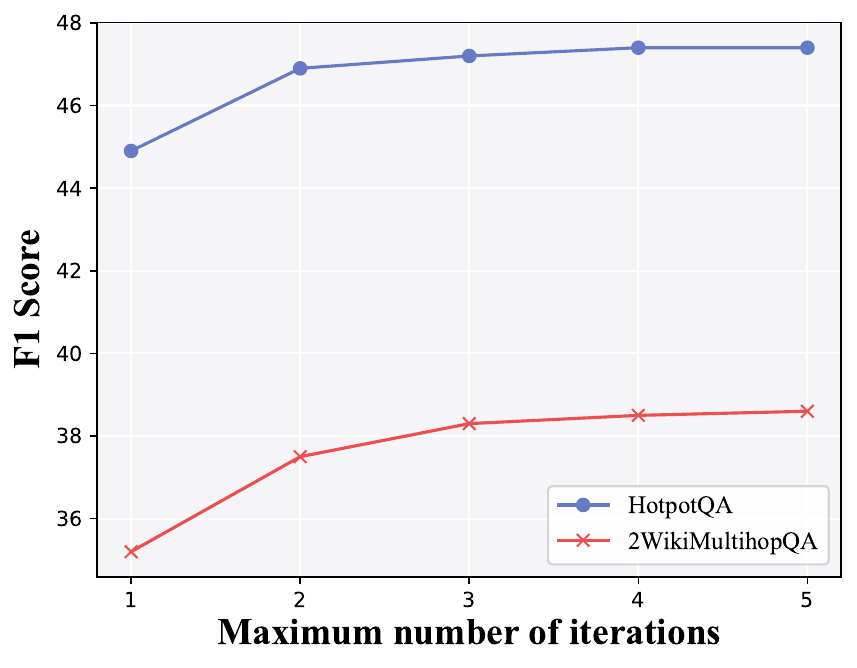}
    \caption{Performance with different maximum number of iterations.}
    \label{fig-number}
\end{figure}
In our main experiments, the maximum number of allowed iterations was set to three, following the construction methods detailed in the HotpotQA and 2WikiMultihopQA papers, where most multi-hop questions are resolved within three hops. However, in practical applications, the maximum number of iterations can significantly affect the performance of iterative RAG methods. Consequently, we conducted comparative experiments by varying the maximum iterations from 1 to 5 to observe the resulting performance changes in ReSP.

As illustrated in Figure \ref{fig-number}, a significant decline in ReSP's performance is observed when the maximum number of iterations is less than three, indicating that gathering information over multiple rounds can enhance response accuracy. However, when the maximum number of iterations exceeds three, the performance improvements become negligible. This is attributed to ReSP's optimization for redundant planning, enabling problem resolution within a reasonable number of hops. These findings are consistent with the average iteration statistics presented in the Section \ref{sec:main}.

\subsection{Case Study}
We present evidence of ReSP addressing over-planning and repetitive planning by comparing cases of IRCoT and ReSP on two questions, as shown in Table \ref{tab:case_study}.

In the first case, the retrieved documents remain consistent because the initial retrieval question is the same. Despite this consistency, the two models take different approaches to their subsequent actions. IRCoT, which combines information processing and planning in a single step, tackles a higher level of task complexity. Unfortunately, it fails to capture information related to \texttt{"Twisted Fortune"}, which leads the model to conclude that additional retrieval is necessary, thereby leading to over-planning and unnecessary steps. In contrast, ReSP effectively and comprehensively gathers the necessary supporting facts related to the overarching question by utilizing the summarizer. As a result, the reasoner accurately assesses that the question can be answered with the information at hand. Consequently, the generator can produce the correct response, allowing the process to conclude successfully after just the first round of retrieval, without the need for further iterations.

In the second case, after the first round of retrieval, both models arrive at similar decisions due to the absence of information concerning the main subject, \texttt{"Rachelle Amy Beinart"} in the retrieved documents. As a result, both models initiate queries for documents specifically related to \texttt{"Rachelle Amy Beinart"}. However, limitations in document coverage or the capabilities of the retriever result in no relevant documents being found on \texttt{"Rachelle Amy Beinart"}. At this juncture, the models adopt different approaches in formulating sub-questions. Without a recorded retrieval trajectory, IRCoT is constrained to make decisions solely based on the current set of information, which leads it to continue querying \texttt{"Rachelle Amy Beinart"}, resulting in repetitive and unproductive planning. In contrast, ReSP employs a strategy that avoids generating previously retrieved sub-questions. This allows it to adapt by adjusting the retrieval subject and crafting a new sub-question focused on \texttt{"film that follows a group of rebels on a mission"}. This shift in focus helps ReSP circumvent repetitive planning, enhancing the efficiency and effectiveness of the retrieval process.


\section{Conclusion}
In this work, we propose an iterative RAG approach that incorporates query-focused summarization. By employing a dual-function summarizer to simultaneously compress information from retrieved documents targeting the overarching question and the current sub-question, we address the context overload and redundant planning issues commonly encountered in multi-hop question answering. Experimental results demonstrate that our method significantly outperforms other single-round and iterative RAG methods. Furthermore, we hope that our empirical analysis will aid the community in practical applications.

\clearpage
\bibliographystyle{ACM-Reference-Format}
\bibliography{sample-base}

\appendix

\end{document}